\documentclass[sn-mathphys]{sn-jnl}
\usepackage{float}

\jyear{2021}%

\theoremstyle{thmstyleone}%
\theoremstyle{thmstyletwo}%

\theoremstyle{thmstylethree}%

\begin{document}

\title{OTB-morph: One-Time Biometrics via Morphing}


\author[1]{\fnm{Mahdi} \sur{Ghafourian}}

\author[1]{\fnm{Julian} \sur{Fierrez}}

\author[1]{\fnm{Ruben} \sur{Vera-Rodriguez}}
\author[1]{\fnm{Aythami} \sur{Morales}}

\author[1]{\fnm{Ignacio} \sur{Serna}}

\affil[1]{\orgdiv{Biometrics and Data Pattern Analytics Lab, BiDA Lab}, \orgname{Universidad Autonoma de Madrid}, \orgaddress{\city{Madrid} \postcode{28049},  \country{Spain}}}



\abstract{Cancelable biometrics are a group of techniques to transform the input biometric to an irreversible feature intentionally using a transformation function and usually a key in order to provide security and privacy in biometric recognition systems. This transformation is repeatable enabling subsequent biometric comparisons. This paper is introducing a new idea to exploit as a transformation function for cancelable biometrics aimed at protecting the templates against iterative optimization attacks. Our proposed scheme is based on time-varying keys (random biometrics in our case) and morphing transformations. An experimental implementation of the proposed scheme is given for face biometrics. The results confirm that the proposed approach is able to withstand against leakage attacks while improving the recognition performance.}




\maketitle

\section{Introduction}\label{sec1}
Biometrics are unique methods of identifying people based on their biological and behavioral characteristics. The advantage of biometric recognition in authentication systems compared to conventional methods such as using passwords or smart cards, has resulted in attracting much attention to this field. However, the widespread usage of biometrics has raised serious security and privacy concerns \cite{2021_EncCrypto_BioSec_Fierrez,19jaswal2021ai}. In addition, standard cryptographic approaches failed to address these concerns due to the noisy nature of biometrics \cite{Freire2007_ICB}. Therefore, a new class of protection methods called biometric template protection (BTP) has emerged as a solution \cite{2017_Access_HEmultiDTW_Marta,2017_PR_multiBtpHE_marta,7192825, Ref8}. Biometric template protection is a set of techniques to preserve the security and privacy of the subject's acquired biometric features. The main goal is to generate a protected biometric reference out of original biometric data which guarantees desired attributes: noninvertibility (irreversibility), revocability (renewability), and unlinkability (nonlinkability) without degrading the recognition performance. Noninvertibility refers to the computational difficulty of obtaining the original biometric template from someone's protected biometric reference. Revocability refers to the ability to change the biometric reference (template) for the same raw input biometric data without affecting the system performance. Unlinkability refers to the computational difficulty of ascertaining the subject's identity by linking multiple biometric references of him. To this end, BTP methods have been introduced and commonly divided into three categories: cancelable biometrics, biometric cryptosystems, and biometrics in encrypted domains \cite{1patel2015cancelable}.

Among these methods, cancelable biometrics \cite{Ref2} is very promising due to their unique features such as providing revocability in case of leakage reports. In general, cancelable biometrics refers to a group of template protection techniques with the primary aim of improving template security and privacy by transforming the original feature using an irreversible transformation function such that the recognition still can be performed but in the transformed domain. These methods should maintain four characteristics for the transformed feature: Diversity, Revocability, Non-invertibility, and Recognition performance. During enrollment in a biometric verification scenario, some biometric data are extracted upon presentation, then the corresponding cancelable biometric transformation is applied to these features (mainly by using auxiliary data) and finally, the result (transformed template) is stored on the server’s database. During verification, when the client presents her biometric feature, the transformed template is extracted similarly to the enrollment phase but by applying the previously stored or known auxiliary data. Lastly, the matching is taken place between the generated cancelable template at the verification phase and the one stored at the enrollment phase called the reference. A general taxonomy of all Cancelable Biometric methods containing six major categories is proposed recently in \cite{2kumar2020cancelable}. 

In the present paper, we adopted the concept of the one-time-pad method \cite{doi:10.1080/01611194.2011.583711} to derive one-time biometrics as a new cancelable biometrics method. The core elements of our proposed scheme are: (i) to use as time-variant keys biometric data generated randomly with natural appearance \cite{2020_JSTSP_GANprintR_Neves}, (ii) combining these keys (random biometrics) with real input biometric data using image/signal morphing techniques \cite{18scherhag2019face}, and (iii) keeping track of the key/template variations in time in a specific secure exchange protocol to enable biometric comparisons while protecting against potential attacks.

The present paper is the extended version of our preliminary research \cite{11OTB-Morph}. In this paper, we extend our previous results by experimenting in a wider range of settings. In particular, first, we increased the number of verification sessions to demonstrate the superiority of the proposed method against iterative optimization attacks in longer runs compared to other protection methods. Second, we used another dataset, Face Mask Lite, a GAN-generated face image dataset, as random biometrics in addition to face images taken from the LFW dataset in order to produce our morphed templates. Third, we used the pre-trained ArcFace and AdaFace models on top of Resnet-50 to report the result of the proposed method in a wider range of face recognition systems. 

The rest of this paper is organized as follows: Sect. 2 summarizes related works in cancelable biometrics. Sect. 3 describes the threat model under which we have conducted our experiments and compared the security improvement of our proposed method with other scenarios. Sect. 4 describes our proposed cancelable biometrics method called OTB-morph. The experimental results for implementing the proposed method on face biometrics and its advantages compared to existing methods are reported in Sect. 5. Finally Sect. 6 concludes the paper.

\section{Related works}\label{sec2}

Over the past two decades, many cancelable biometrics research has been carried out due to the increasing usage of biometric-based authentication. In this section, we provide a brief description of most noticeable attempts in this area. 

The concept of cancelable biometrics was first introduced in \cite{3ratha2001enhancing} to enhance security and privacy in biometric-based authentication systems. Among early noticeable attempts, Jin et al. \cite{4jin2004biohashing} proposed a Random projection-based technique called BioHashing. This method projects biometric features to a random space by taking the inner product between a tokenized pseudo-random number and the subject's fingerprint. In 2005, Ang et al. \cite{5ang2005cancelable} proposed a key-dependent cancelable template where a geometric transformation was applied to features extracted from a fingerprint so as to protect minutiae templates. In 2006, Chin et al. \cite{6chin2006high} presented a work securing iris features coined as S-Iris encoding. To this end, they iterated inner products between secret pseudo-random numbers and the iris features. In 2007, the first alignment-free cancelable biometrics was introduced by Lee et al. \cite{7lee2007alignment}. They protected fingerprint templates by extracting rotational and translational invariant features from each minutia. Later that year, Ratha et al. \cite{8ratha2007generating} suggested three different methods (Cartesian, Polar, and Surface Folding) to transform minutia positions extracted from a fingerprint image. These transformations were aimed at distorting original biometrics and offering noninvertibility and revokability. However, soon after Quan et al. \cite{9quan2008cracking} showed that most of the transformed minutia in \cite{8ratha2007generating} could be exactly inversed. 

More recently Maiorana et al. \cite{10maiorana2010cancelable} proposed a convolution-based noninvertible transformation named BioConvolving, which can be applied to any sequence-based biometric. They practiced their approach on online signature biometrics and its security relies on the difficulty of solving a blind deconvolution problem. The same year, Ouda et al. \cite{11ouda2010tokenless} proposed a cancelable biometric scheme for protecting Iris-Codes. Their method extracts consistent bits from Iris-Codes and further encodes them using a random encoding process referred to as BioEncoding. Another research \cite{12pillai2010sectored} generated cancelable iris biometrics using sectored random projections that year. This method mitigates the performance degradation due to eyelids and eyelashes. In 2012, Ferrara et al. \cite{13ferrara2012noninvertible} provided noninvertibility based on dimensionality reduction and binarization to protect Minutia-Cylinder-Code, which is a local minutia representation. Later, Gomez-Barrero et al. \cite{2016_InformationSciences_Marta,marta14FaceBF,rathgeb15IWBF_IrisFaceBF} proposed an alignment-free cancelable iris template based on Bloom filters. They argued that successive mapping of parts of a binary biometric template to a Bloom filter represents a noninvertible transformation. Chin et al. \cite{15chin2014integrated} proposed another template protection technique in 2014 by fusing fingerprint and palmprint at the feature level using client-specific keys. Three years later, Lai et al. \cite{16lai2017cancellable} introduced a cancelable iris template generation method coined as Indexing-First-One (IFO) hashing. The method is inspired by Min-hashing and extended by using modulo threshold functions and P-order Hadamard products. In 2019, Sadhya and Raman \cite{17sadhya2019generation} generated a cancelable iris template using randomized bit sampling. Their method (LSC) is functionally based on the notion of Locality Sensitive Hashing (LSH) in which two items that are relatively close to each other, are hashed into the same location \cite{Freire2007_ICB}. In 2020, Kirchgasser et al. \cite{Ref1} compared cancelable approaches using finger vein biometrics in both the signal and the feature domains. They reported that for most experimental settings, it is possible to track a subject across several instances generated with various keys. In the same year, the same research group reported in \cite{Ref3} that considering state-of-the-art deep-learning methods, warping-based cancellable biometrics is no longer a protection scheme. Next year, Badr et al. \cite{Ref4} presented a cancellable face recognition scheme that is based on face image encryption with Fractional-Order (FO) Lorenz chaotic system. In 2022, Dong et al. \cite{Ref5} proposed a deep learning-based cancellable biometric scheme for face identification (one-to-many matching). In this research they used a Deep Rank Hashing (DRH) network and a randomized lookup table function to transform a raw face image into discriminative yet compact binary face hash codes. Finally, Chang et al. \cite{Ref6} proposed a multi-biometric cancelable approach using fuzzy extractor and a bit-wise encryption scheme to transform a biometric template to a protected template by means of a secret key generated from another biometric template.

What makes our research different from these works, is addressing the iterative optimization attacks. This is very important, because with the appearance of adversarial examples as a branch of iterative optimization, the possibility of this threat has increased tangibly. To the best of authors knowledge, there is no prior biometric template protection method taking into account addressing the threat of iterative optimization attacks.

\section{Threat model}\label{sec3}

\begin{figure*}[tbh]
 \centering 
 \includegraphics[trim={3cm 4cm 2cm 2cm},clip,width=130mm,scale=0.5]{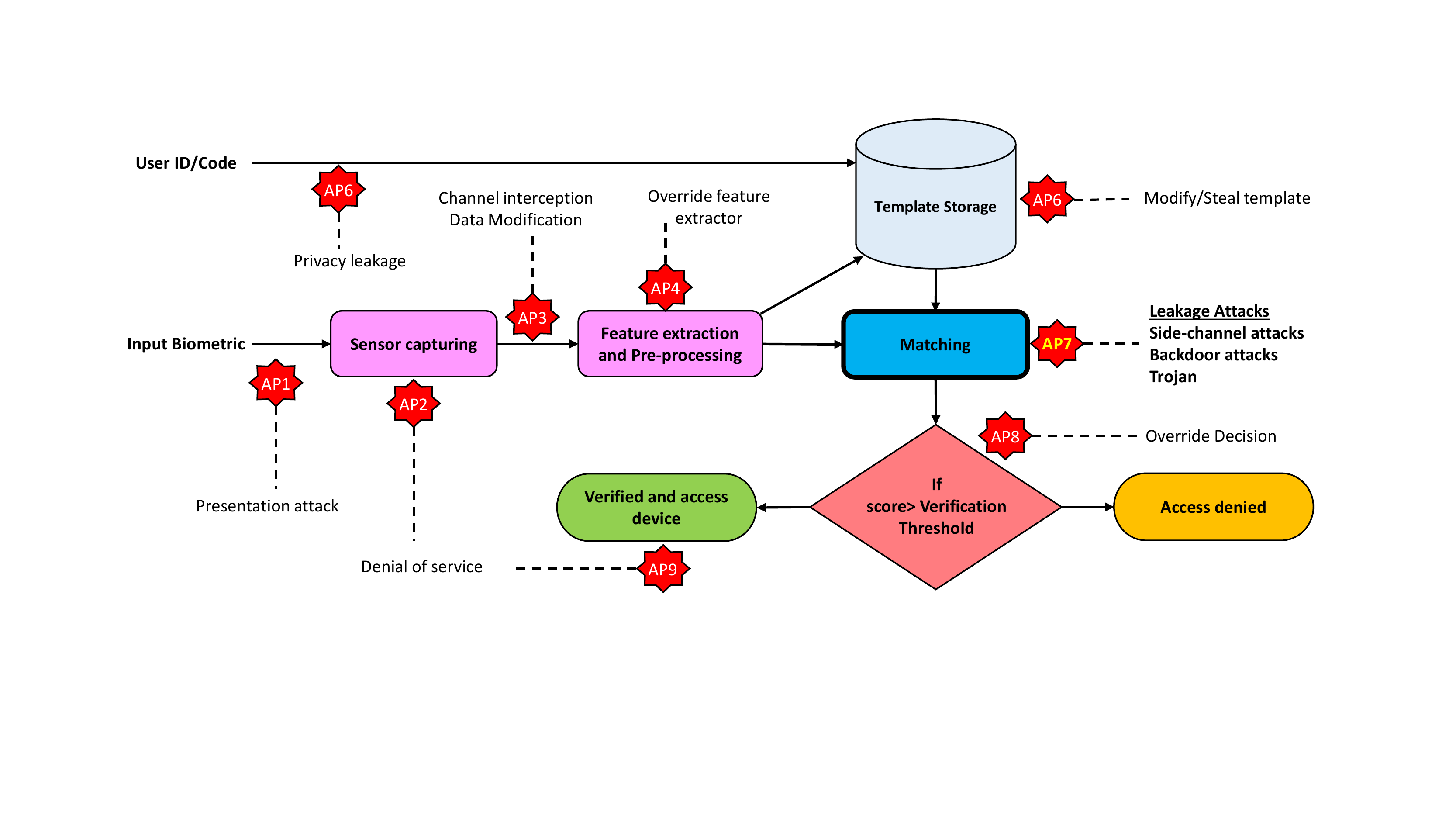}
  \caption{Attack points (AP) in a generic biometric system.}
  \label{fig:AttackPoints}
\end{figure*}

Biometric systems can be the target for an attacker to conduct malicious activities, including impersonation. The possible attack points are positioned in a generic biometric system in Figure~\ref{fig:AttackPoints} \cite{2021_EncCrypto_BioSec_Fierrez,19jaswal2021ai}.

This paper is focused on addressing three challenges: (i) privacy leakages at attack point AP6, (ii) injection attacks at AP4, and (iii) leakage threats at AP7. This threat model specifies the Adversary's goal, capabilities, and knowledge under which the aforementioned attacks are feasible. In particular, we assume that:

\begin{itemize}
 \item The attacker is able to eavesdrop the communication channel from AP6 where genuine clients request verification.
  \item The similarity score of biometric templates at the matching phase is leaked to the attacker through any wide-range means of leakage attacks such as backdoors, trojans, side-channel attacks \cite{GALBALLY2020101902,2009_GalballyBioID}, etc. 
  \item The attacker is able to get the similarity score between an arbitrary biometric input and the feature reference of victims from AP7 for some verification sessions, not necessarily consecutive.
  \item The attacker possesses the knowledge of the underlying model with which the protected template (victim's reference) is generated from the input biometric data (i.e., the biometric feature extractor).
  \item The attacker is able to get at least one biometric input of the victim.
  \item The attacker is able to override the feature extractor and can inject his biometric features in AP4. 
\end{itemize}

Using this leaked score or the obtained biometric input, the attacker can maximize the similarity of his arbitrary input biometric compared to the victim's reference by iterative optimization, an adversarial perturbation that is added to the attacker image after each comparison with the victim image gradually in order to lower the similarity score. e.g., deep leakage from gradient \cite{zhu2020deep}, hill-climbing \cite{Galbally_2009PR,barrero13PRLmultimodalAttack,barrero12ICB}. 

\section{Proposed scheme: OTB-morph}\label{sec4}

\begin{figure*}[tbh]
 \centering 
 \includegraphics[trim={3.5cm 4cm 2cm 2cm},clip,width=130mm,scale=0.5]{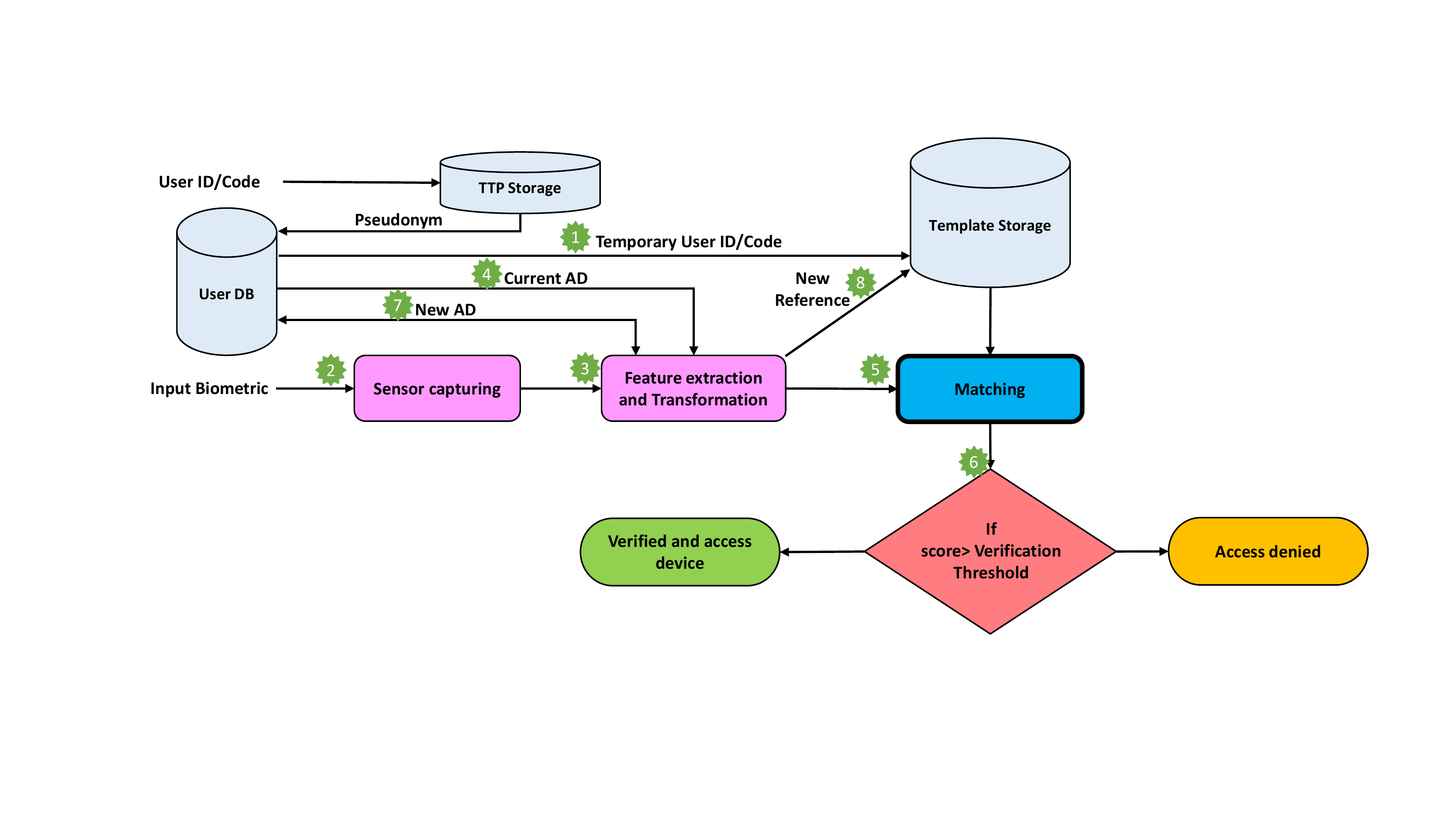}
  \caption{Architecture of the proposed One-Time Biometrics scheme (OTB-morph).}
  \label{fig:Architecture}
\end{figure*}

The aim of the proposed scheme is to address both privacy leakages at attack point 6 (AP6, see Figure~\ref{fig:AttackPoints}) and leakage attacks at attack point 7 (AP7). The block diagram showing the architecture and data flow of the proposed scheme in a generic biometric system is shown in Figure~\ref{fig:Architecture}. 

There are three parties involved during biometric verification. A \textit{Client} who wants to be verified in a \textit{Server} using a temporary identity that has been assigned to him by a \textit{Trusted Third Party (TTP)}. It is assumed that enrollment phases in both server and TTP are already accomplished and the corresponding Auxiliary Data (AD) and Pseudonyms are stored on a secure element in the client’s device or his smartcard (note that the complete process of the proposed method is explained in detail later with an example in face biometrics). In this regard, the client starts the verification session by sending his request to the server using one of his stored pseudonyms (num 1). Pseudonyms are temporary identities that have been assigned to the client previously by the Trusted Third Party (TTP). We adopted the Pseudonym architecture described in Section~\ref{subsec1} from \cite{20ghafoorian2020anonymous} for our problem. Upon receiving the answer from the server, the client presents his biometric to the input sensor (num 2) and the extracted feature will be transformed to a cancelable biometric template (num 3) using the current AD (num 4) that he has stored on his device/smartcard from the enrolment process. In the next step, the produced cancelable biometric template is sent to the server domain to be compared in the biometric matcher with the feature reference of the client (num 5). Depending on the verification threshold, access is granted or denied (num 6). Generally, most cancelable biometrics techniques need Auxiliary Data (AD) to compute the transformation of biometric features. This AD can be a password, a random number, etc., and it is usually permanent until a leakage on the respective cancelable template is reported. In our proposed method, these auxiliary data are random biometrics (e.g., GAN-generated synthetic faces \cite{2020_JSTSP_GANprintR_Neves}, LSTM-generated synthetic handwriting \cite{2021_AAAI_DeepWriteSYN_Tolosana}, etc.), sent to the client inside the pseudonym sets managed by the TTP. When the matching is successful, we propose to re-enroll by picking a new random biometric (AD) (num 7) and combining it with the already extracted feature. The resulting cancelable template is stored as a new reference (num 8) in the server's database. Finally, the new AD is stored on the client's device replacing the previous one. Here with OTB-morph, we propose to combine the random and the input raw biometrics via image- or signal-morphing, depending on the nature of the biometrics at hand.

The notations used in this paper are described in Table~\ref{tab:notation}.

\subsection{Enrollment in Trusted Third Party (TTP)}\label{subsec1}

The client registers in TTP by sending his $ID_{C}$ and his public key $Q_{C}$. Then TTP stores these data and sends back $n$ temporary identities (called pseudonym) $ PN_{C}^{i},i=1,2,...,n$ with his $ID_{TTP}$ to the client. Upon receiving $n$ pseudonym sets, the client stores all of them protected in his device. These pseudonym sets are meant to be used per verification session. The structure of the pseudonym and corresponding signature for client C is as follows:
\begin{equation} \label{eq1}
\begin{split}
 & PN_{C}^{i}= \{TID_{C}^{i} \mathbin\Vert AEnc(Q_{c},RF_{C}^{i}) \mathbin\Vert PID_{TTP} \mathbin\Vert LT_{C}^{i} \mathbin\Vert S_{TTP}^{i}\} 
\end{split}
\end{equation}
\begin{equation} \label{eq2}
\begin{split}
 & S_{TTP}^{i}=Sig(d_{TTP},TID_{C}^{i} \mathbin\Vert AEnc(Q_{c},RF_{C}^{i}) \mathbin\Vert PID_{TTP} \mathbin\Vert LT_{C}^{i} \mathbin\Vert S_{TTP}^{i}) 
\end{split}
\end{equation}
\begin{table}
  \caption{Notations used in this paper}
  \label{tab:notation}
  \begin{tabular}{ll}
    \toprule
    Notation&Description\\
    \midrule
    $N_{x}$&The nonce generated by the party x\\
    $d_{x}$&The private key of the party x\\
    $Q_{x}$&The public key of the party x\\
    $PN_{C}^{i}$&The $i^{th}$ pseudonym set of client\\
    $LT_{C}^{i}$&The lifetime of $i^{th}$ pseudonym set of client\\
    $ID_{x}$&The identifier of the party x in the transport protocol\\
    $PID_{TTP}$&The permanent identity of TTP\\
    $TID_{C}^{i}$&The $i^{th}$ temporary identity of client\\
    $SK_{SC}^{i}$&The $i^{th}$ pseudonym set shared secret key\\
    $RF_{C}^{i}$&The $i^{th}$ randomly created face image using as AD\\
    $F_{C}^{i}$&Client presented face at $i^{th}$ session\\
    $R_{C}^{i}$&Client feature reference at $i^{th}$ session\\
    $MF_{C}^{i}$&The morphed face of client used at $i^{th}$ session\\
    $MGF$&The morph generation function\\
    $Mtch$&The face matching function\\
    $KGF$&The symmetric key generation function\\
    $AEnc(k,m)$&\vtop{\hbox{\strut Asymmetric encryption of the message m}\hbox{\strut  with the key k}}\\
    $Enc(k,m)$&\vtop{\hbox{\strut Symmetric encryption of the message m}\hbox{\strut  with the key k}}\\
    $Sig(k,m)$&Signing the message m with the key k\\
    $resp_{x}$&The response of party x\\
    $\mathbin\Vert$&The concatenation operation\\
  \bottomrule
\end{tabular}
\end{table}

\subsection{Enrollment in Server}\label{subsec2}
The genuine client enrolls in the server by presenting his face. Upon this, the system picks a random pseudonym and applies a random face image as auxiliary data to the cancelable method. This face image is an arbitrary face image (real or artificial) that is not repeated in any pseudonym sets before or in the future. Then, a face morphing transformation is applied to both face images to generate the protected template. Next, the cancelable template is stored on the server's database as the client's biometric reference. Finally, the arbitrary face extracted earlier from the pseudonym set is recorded as the current auxiliary data (current AD) in a secure element at the client's device and the corresponding pseudonym is discarded. The process of client C registering in the server through a secure channel is described in the following steps:

\textbf{Step-1:} The client presents his face $F_{C}^{i}$  and picks a random pseudonym set from his storage, extracts $RF_{C}^{i}$ by doing $ADec(d_{C},AEnc(Q_{C},RF_{C}^{i}))$  and computes $MF_{C}^{i}=MGF(F_{C}^{i},RF_{C}^{i})$, then sends the message $M_{1}=\{ID_{C},MF_{C}^{i},PN_{C}^{i}\}$ to the server to request registration.

\textbf{Step-2:} Upon receiving $M_{1}$, server first checks whether $LT_{C}^{i}$ is valid. If it does not hold, the server terminates the registration; otherwise, it tries to verify the authenticity of $PN_{C}^{i}$ by decrypting $S_{TTP}^{i}$ in $PN_{C}^{i}$ using $Q_{TSM}$ and compares the obtained parameters with the corresponding ones existing in the pseudonym content. If this authenticity does not hold, the server terminates the session; otherwise, it generates a random secret $SK_{SC}^{i}$ corresponding to the client's $i^{th}$ pseudonym set. Then, server stores $PN_{C}^{i}$, $MF_{C}^{i}$ and $SK_{SC}^{i}$ for the client's temporary identity $TID_{C}^{i}$ on its database. Finally, it sends the message $M_2=\{ID_{S},N_{S},SK_{SC}^{i}\}$ to the client. Henceforth, we call the $MF_{C}^{i}$, the one that the server stores on its database, client's reference $R_{C}^{i}$.

\textbf{Step-3:} Upon receiving the message $M_{2}$, the client stores $\{TID_{C}^{i},RF_{C}^{i},SK_{SC}^{i}\}$ on his device protected as the current credentials and drops $PN_{C}^{i}$.

\subsection{Verification protocol using the proposed method}\label{subsec3}
\begin{figure*}[htp]
\centering
  \includegraphics[trim={4cm 5cm 3cm 3cm},clip,width=1.3\linewidth]{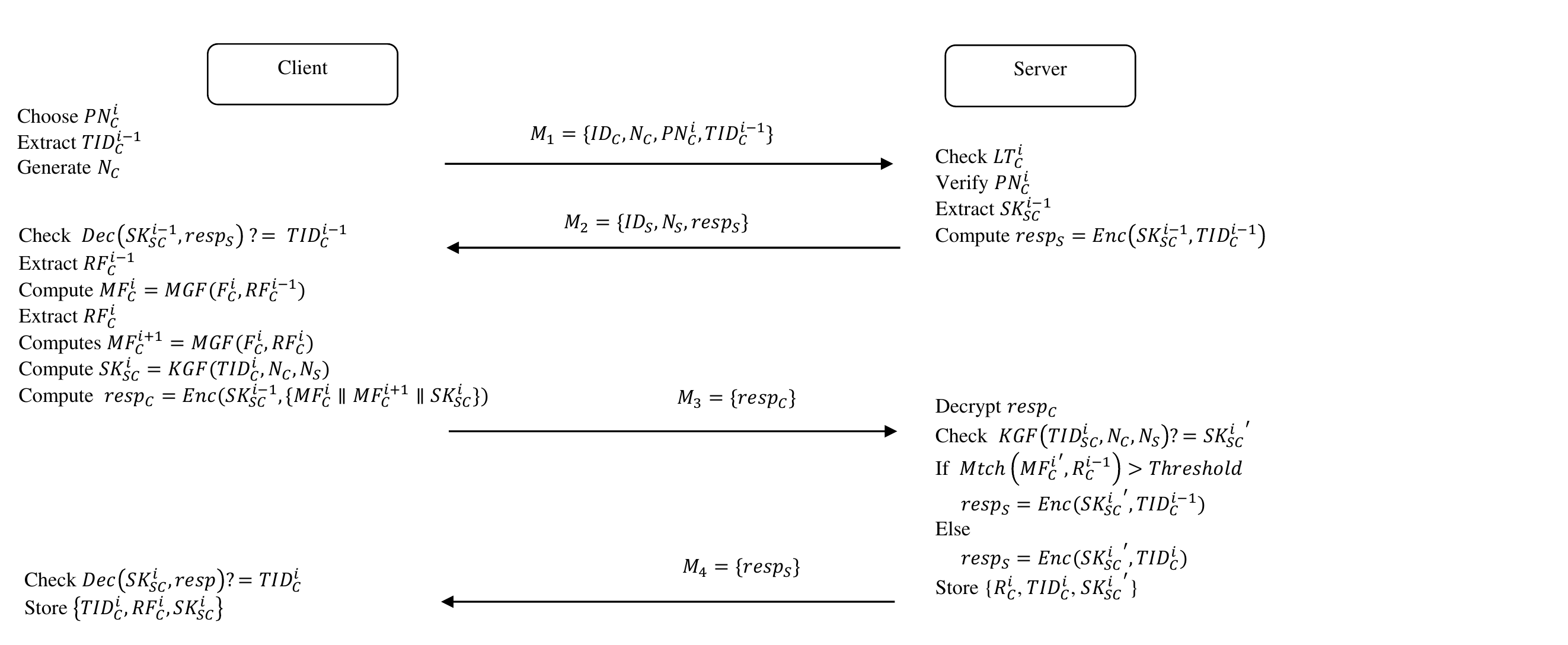}
  \caption{Biometric verification protocol using the proposed method}
  \label{fig:Protocol}
\end{figure*}

In order to establish a face verification between client and server using the proposed method, the following steps are provided. The summary of these steps is given in Figure~\ref{fig:Protocol}.

\textbf{Step-1:} Client starts the session by picking a random pseudonym $PN_{C}^{i}$, extracting $TID_{C}^{(i-1)}$ from his storage, and generating a nonce $N_{C}$. Then, he sends the message $M_{1}=\{ID_{C},N_{C},PN_{C}^{i},TID_{C}^{(i-1)}\}$ to the server.

\textbf{Step-2:} Upon receiving $M_{1}$, server first checks whether $LT_{C}^{i}$ is valid. If it does not hold, the server terminates the session; otherwise, it tries to verify the authenticity of $PN_{C}^{i}$ by decrypting $S_{TTP}^{i}$ in $PN_{C}^{i}$ using $Q_{TSM}$ and compares obtained parameters with the corresponding ones existing in the pseudonym content. If this authenticity does not hold, server terminates the session; otherwise, server extracts $SK_{SC}^{(i-1)}$ corresponding to $TID_{C}^{(i-1)}$ from its database, then computes $resp_{S}=Enc(SK_{SC}^{(i-1)},TID_{C}^{(i-1)})$ by doing symmetric encryption. Finally, server generates a nonce and replies to client by sending the message $M_{2}=\{ID_{S},N_{S},resp_{S}\}$.

\textbf{Step-3:} Upon receiving the message $M_{2}$, the client verifies $resp_{S}$ by checking whether the equation \ref{eq3} holds. If it does not hold, client terminates the session and starts a new one; otherwise, he presents his face $F_{C}^{i}$ to his device's camera, extracts $RF_{C}^{(i-1)}$ from his storage and computes $MF_{C}^{i}=MGF(F_{C}^{i},RF_{C}^{(i-1)})$. Next, he extracts $RF_{C}^{i}$ from $PN_{C}^{i}$ by doing $ADec(d_{C},AEnc(Q_{C},RF_{C}^{i} ))$ and computes $MF_{C}^{(i+1)}=MGF(F_{C}^{i},RF_{C}^{i})$. Finally, client computes $SK_{SC}^{i}=KGF(TID_{C}^{i},N_{C},N_{S})$ and equation \ref{eq4}, then he sends the message $M_{3}=\{resp_{C}\}$ to the server.

\begin{equation} \label{eq3}
    Dec(SK_{SC}^{(i-1)},resp_{S})=TID_{C}^{(i-1)}
\end{equation}
\begin{equation} \label{eq4}
\begin{split}
 & resp_{C}= Enc(SK_{SC}^{(i-1)},\{MF_{C}^{i} \mathbin\Vert MF_{C}^{(i+1)} \mathbin\Vert SK_{SC}^{i}\})
\end{split}
\end{equation}

\textbf{Step-4:}  Upon receiving the message $M_{3}$, server decrypts $resp_{C}$ using $SK_{SC}^{(i-1)}$ and checks whether the equation \ref{eq5} holds. If it does not hold, server terminates the session; otherwise, it computes $Mtch({MF_{C}^{i}}^{'},R_{C}^{(i-1)})$. If the corresponding result is not above the face matching threshold, server computes $resp_{S}=Enc({SK_{SC}^{i}}^{'},TID_{C}^{(i-1)})$ and sends the message $M_{4}=\{resp_{S}\}$ to the client and terminates the session; otherwise, server first drops $R_{C}^{(i-1)}, TID_{C}^{(i-1)}, SK_{SC}^{(i-1)}$ and replace them with ${MFC_{C}^{(i+1)}}^{'}$ as $R_{C}^{i}, TID_{C}^{i}, {SK_{SC}^{i}}^{'}$ respectively. Then, it computes $resp_{S}=Enc({SK_{SC}^{i}}^{'},TID_{C}^{i})$ and sends the message $M_{4}=\{resp_{S}\}$ to the client.

\begin{equation} \label{eq5}
\begin{split}
    KGF(TID_{SC}^{i},N_{C},N_{S})={SK_{SC}^{i}}^{'}
\end{split}
\end{equation}

\textbf{Step-5:} Upon receiving the message $M_{4}$, client checks whether $Dec(SK_{SC}^{i},resp_{S})=TID_{C}^{i}$ holds. If it does not, he repeats Step-3 from face presentation part; otherwise, he drops previously stored credentials and replaces them with $\{TID_{C}^{i},RF_{C}^{i},SK_{SC}^{i}\}$.

\begin{figure*}[h]
 \centering 
 \includegraphics[trim={2cm 0cm 0 0cm},clip,width=130mm,scale=0.5]{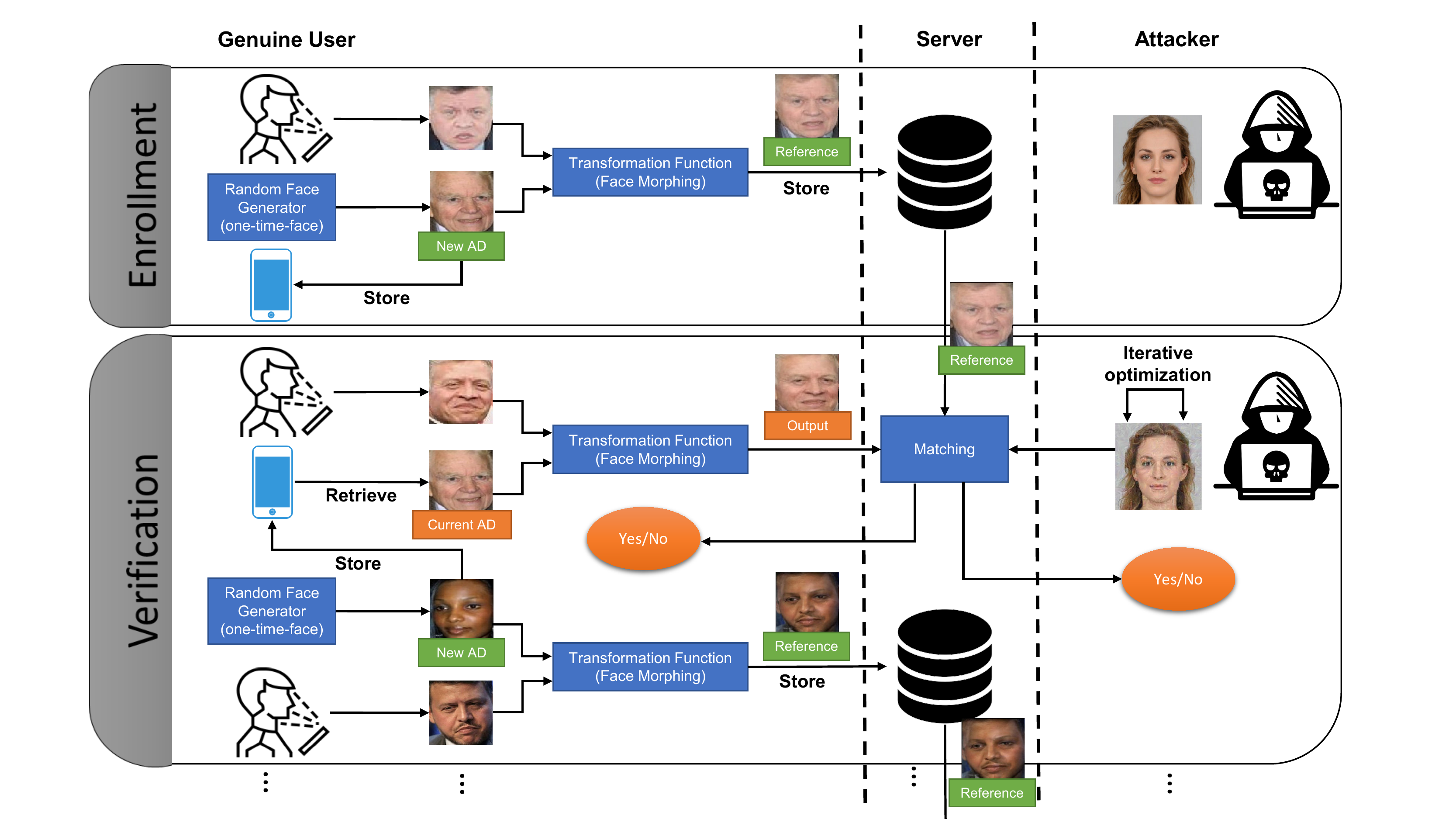}
 \caption{Visual examples of the process of the proposed One-Time Biometrics via Morphing (OTB-morph) for enrollment and various verification sessions (genuine clients and attackers).}
 \label{fig:process}
\end{figure*}

For a better understanding of readers, the whole operation of the proposed cancelable biometrics method including client enrollment and verification is depicted in Figure~\ref{fig:process}.

\section{Experiments}\label{sec4}
\begin{figure*}[tbh]
 \centering 
 \includegraphics[trim={2cm 6cm 0 6cm},clip,width=125mm,scale=0.5]{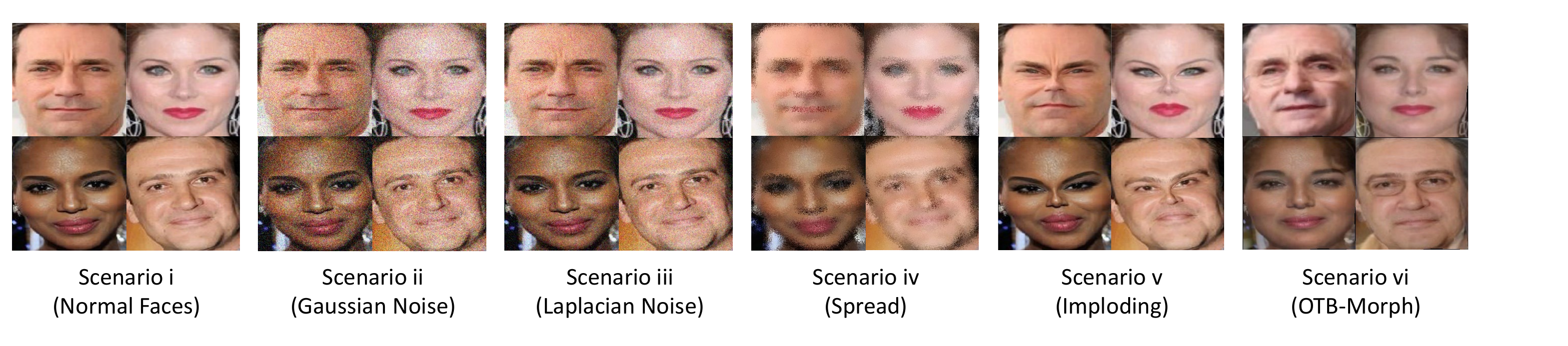}
 \caption{Examples of experimented scenarios }
 \label{fig:scenarios}
\end{figure*}

In this paper, we implemented an attack framework using iterative optimization in which the adversary who obtained the matching score of the victim's biometric feature explained in section 3, is able to update an arbitrary face image such that the corresponding score (Euclidean Distance in our experiments, therefore dissimilarity score) of it with respect to the victim's reference becomes lower than the verification threshold \cite{Galbally_2009PR}. In other words, using this attack framework, the adversary is able to manipulate his arbitrary face image and successfully impersonate a legal client. In order to confirm the weakness of current cancelable biometric methods against leakage attacks, we implemented our experiments with respect to seven scenarios: (i) Face verification without applying any protection method; (ii) Face verification protected by applying Gaussian noise as cancelable transformation; (iii) Face verification protected by applying Laplacian noise as cancelable transformation; (v) Face verification protected by applying spread, a transformation that replaces each pixel with a random pixel value found in a radius nearby; (v) Face verification protected by applying imploding, a transformation that pulls pixels into the middle of the image; (vi-a) Face verification protected by applying the proposed method using the LFW dataset as random biometric; and (vi-b) Face verification protected by applying the proposed method using the Face Mask Lite dataset as random biometric. An example showing the input biometric of the experimented scenarios is depicted in Figure~\ref{fig:scenarios}. The experiments are conducted using the following face datasets: on the one hand, VGGFace2 \cite{21cao2018vggface2} and CASIA \cite{22yi2014learning} as genuine client's biometrics, and on the other hand, LFW \cite{2018_TIFS_SoftWildAnno_Sosa,24huang2008labeled} and Face Mask Lite\footnote{https://www.kaggle.com/datasets/prasoonkottarathil/face-mask-lite-dataset} to select the random face images for the morphing operations.

\subsection{Implementation details}

We performed our implementation on pre-trained Resnet-50 \cite{23he2016deep}, pre-trained ArcFace \cite{24deng2019arcface} and pre-trained AdaFace \cite{Ref7}, CNN models proposed for general image recognition tasks using two groups of datasets. As the first group, we used VGGFace2 \cite{21cao2018vggface2} and Casia \cite{22yi2014learning} datasets, two face datasets that contain multiple faces of the same individual. The images in these datasets are utilized as probe faces of genuine clients during verification sessions. Regarding the second group, we used LFW \cite{2018_TIFS_SoftWildAnno_Sosa,24huang2008labeled} and Face Mask Lite (we used face images without masks) as the auxiliary data (a random seed) to create morph faces for our proposed OTB-morph scheme. In other words, our method takes two input faces, one from the first group as the probe biometric feature of the subject meant to be protected, and the second input is a randomly chosen face image from the second group to be morphed with the first image. 

\subsubsection{Image morphing}
Image morphing is an image processing technique that can transform one image into another image. Applied to face images, morphing is being used to generate artificial faces which resemble the biometric characteristics of at least two input individuals in image and feature space \cite{18scherhag2019face}. Morphed faces can be generated using various methods from simple image overlaying to Generative Adversarial Networks (GAN). The most popular morphing method is landmark-based, which consists of three steps: (i) determining a correspondence between the two contributing face images; (ii) warping, which means distorting both features such that the corresponding facial elements (e.g., eye, nose, mouth) are geometrically aligned; and (iii) blending, which refers to the process of merging the color values of wrapped images. In our experiments, we use landmark-based morphing as a transformation function for our proposed cancelable biometrics method. There are many landmark detection algorithms such as \cite{Ref9} for face biometrics. Our morphing implementation is based on Dlib for landmark detection \cite{18scherhag2019face} and OpenCV for image processing \cite{26scherhag2020morphing}.

The landmarks locations obtained from both face images are warped by averaging the pixel positions. After moving the pixels we apply image warping based on Delaunay triangulation \cite{25venkatesh2021face}. Our morphing method has a parameter $\alpha$ between 0 and 1 that trades off the contribution of each input image: smaller $\alpha$ generates an output more similar to the first contributed face image (probe face in our case), and higher $\alpha$ results in a morphed face more alike to the second contributed face image (random face). In these experiments, we selected $\alpha=0.5$ to keep the trade-off.

\subsection{Performance and security metrics}
We use the Equal Error Rate (EER) to evaluate and compare the verification performance of our proposed method with other scenarios. EER is the point where the False Acceptance Rate (FAR) and False Rejection Rate (FRR) are equal, where FAR is the percent of unauthorized clients (random impostors\footnote{This kind of impostors are different to the attackers considered in Section~\ref{sec3}, who have much more information to attack the system compared to a random impostor that just tries to illegally access the system by using his own face input and no other methods to improve the attack success.}) incorrectly verified as a valid client (genuine) while FRR is the percent of incorrectly rejected valid clients. The evaluation metric EER describes the overall accuracy of a biometric system. In general, the lower the EER value, the higher the accuracy of the biometric system. 

Regarding security evaluation, the vulnerability of the compared cancelable biometrics schemes under the considered Threat Model (cf. Section~\ref{sec3}) is analyzed by looking at the capability of the attacker to minimize the dissimilarity score of his arbitrary face image by iterative optimization exploiting the leaked matching score. More specifically, we measure the Attack Success Rate (ASR) to assess and compare the vulnerability of all experimental scenarios \cite{Galbally_2009PR}.

\subsection{Results}

\begin{figure*}[!tbh]
 \centering 
 \includegraphics[trim={0cm 1.5cm 0 1.2cm},clip,width=118mm,scale=0.5]{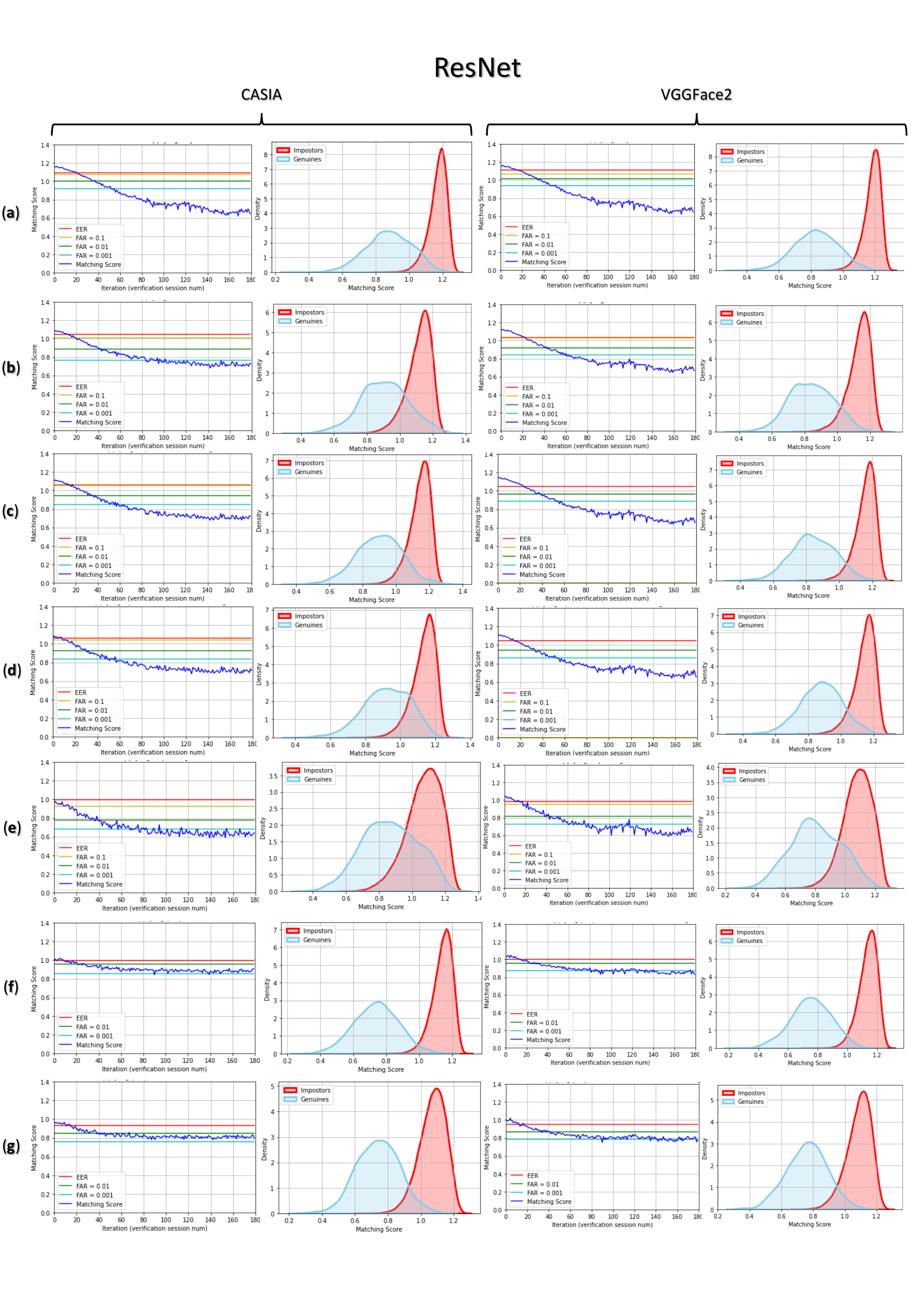}
  \caption{Comparison of practiced scenarios on the ResNet model: The first column is attacking matching (dissimilarity) score evolution on the CASIA dataset (positioned on top of decision thresholds at EER and various FAR). The second column is the genuine and random impostor distributions of the seven considered cancelable biometrics approaches on the CASIA dataset corresponding to different rows. Rows represent different scenarios: (a) without applying cancelable biometrics; (b) applying Gaussian noise; (c) applying Laplacian moise; (d) applying spreading (e) applying imploding; (f) applying the proposed OTB-morph scheme using the LFW dataset as random biometric; and (g) applying the proposed OTB-morph scheme using Face Mask Lite dataset as random biometric. Third and fourth columns: idem on VGGFace2 dataset.}
  \label{fig:resultsResNet}
\end{figure*}

\begin{figure*}[!tbh]
 \centering 
 \includegraphics[trim={0cm 1.5cm 0 1cm},clip,width=120mm,scale=0.5]{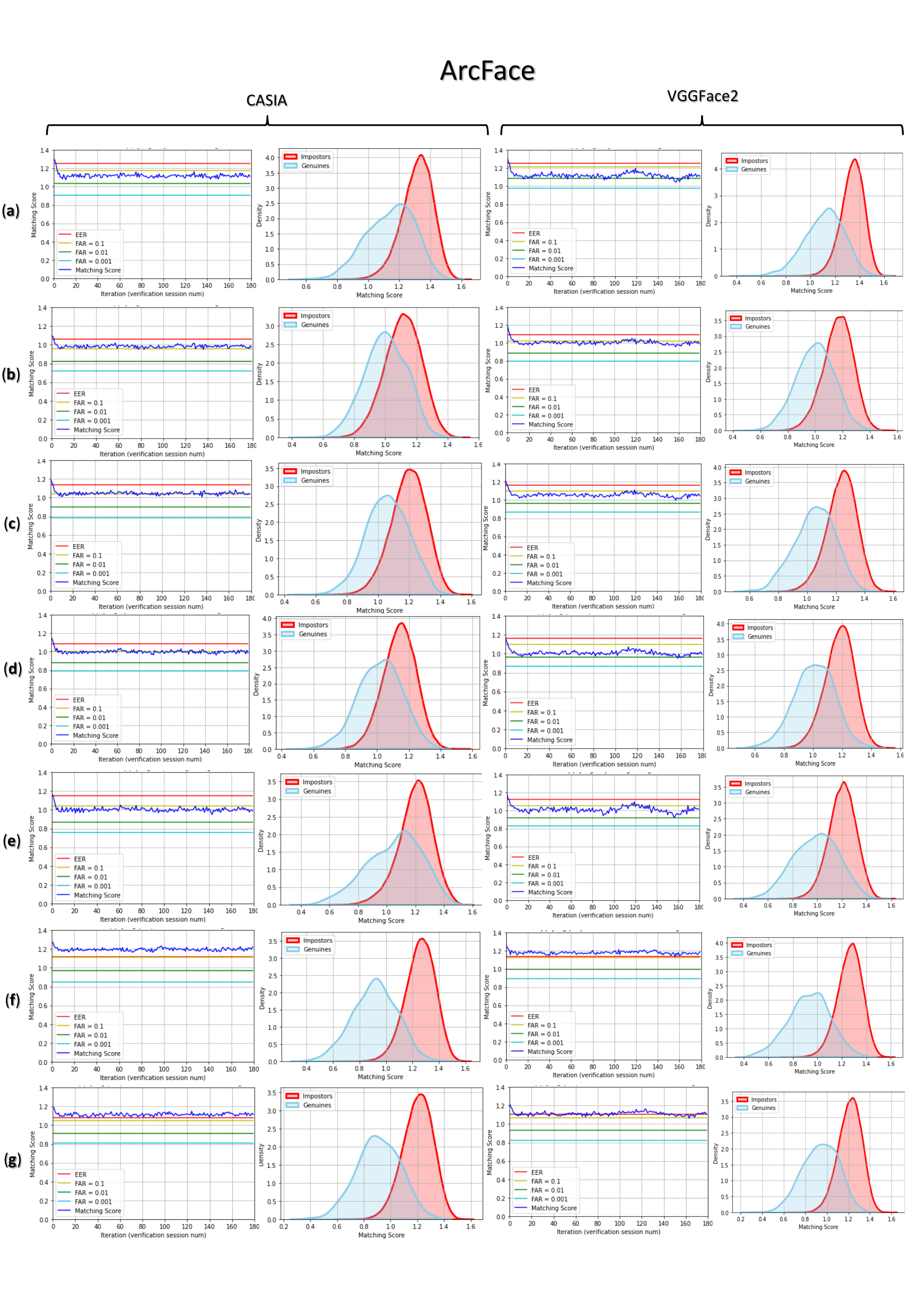}
  \caption{Comparison of practiced scenarios on ArcFace model: descriptions are the same as the caption in Figure~\ref{fig:resultsResNet}.}
  \label{fig:resultsArcFace}
\end{figure*}

\begin{figure*}[!b]
 \centering 
 \includegraphics[trim={0cm 1.5cm 0 1cm},clip,width=120mm,scale=0.5]{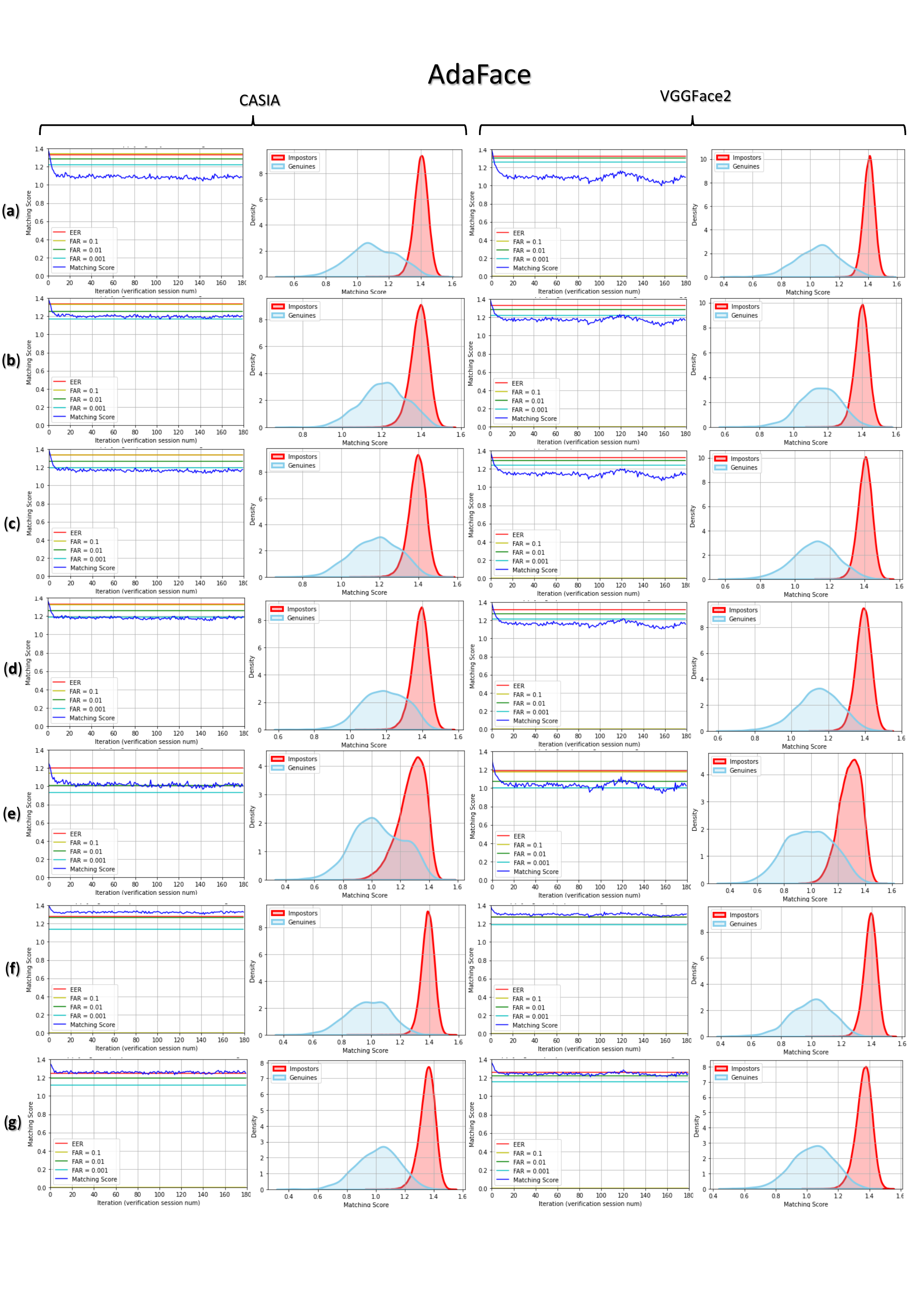}
  \caption{Comparison of practiced scenarios on AdaFace model: descriptions are the same as the caption in Figure~\ref{fig:resultsResNet}.}
  \label{fig:resultsAdaFace}
\end{figure*}

The results of our experiments on Resnet-50, ArcFace, and AdaFace models are demonstrated in Figures~\ref{fig:resultsResNet}, ~\ref{fig:resultsArcFace} and ~\ref{fig:resultsAdaFace} respectively. In general, these three figures consist of seven rows, each of them representing one of the seven scenarios we implemented: (a) not applying any protection method; (b)  applying Gaussian noise; (c) applying Laplacian noise; (d) applying spread;(e) applying imploding; (f) applying our proposed method OTB-morph using the LFW dataset as random biometric; and (g) applying our proposed method OTB-morph using face mask lite dataset as random biometric. These Figures also comprise four columns: the first one shows the attacking matching (dissimilarity) score evolution on the CASIA dataset. The second column shows the score distributions obtained for the seven scenarios considered with respect to the CASIA dataset. The last two columns are similar to the first two columns but with respect to the VGGFace2 dataset. In the plots representing attacking matching (dissimilarity) score evolution (columns 1 and 3), in the vertical axis we can see multiple horizontal lines representing the decision threshold location at the EER point and various FAR points (see the figure legends). Additionally, these plots represent the time evolution of the attacking score in 180 consecutive iterations, which we call verification sessions (from left to right in each plot). 

The scenarios that we used as transformation functions differ from each other in terms of the type of perturbation they apply to the input image. While some scenarios simply add different types of noises, others change the structure of images. Therefore, to compare the proportion of perturbations applied to input images in each experimental scenario, we used two full-reference image quality metrics: the Mean Square Error (MSE) and the Structural Similarity Index (SSIM). The outputs of these metrics are reported in Table\ref{tab:results_Resnet}.

\begin{table}[h]
\centering
\caption{The rate of perturbations applied to images in each experimented scenario. MSE stands for the Mean Square Error (the lower the value, the more similar the perturbed image is to the original one), SSIM stands for the Structural Similarity Index value (ranging between 0 to 1 where 1 means a perfect match between the perturbed image and the original}
\label{tab:results_Resnet}
\resizebox{\textwidth}{!}{%
\begin{tabular}{|l|cccccc|}
\hline
\multirow{2}{*}{Metric} & \multicolumn{6}{c|}{Scenarios}                                                                                                        \\ \cline{2-7} 
                        & (ii) Gaussian noise & (iii) Laplacian noise & (iv) Spread transformation & (v) Imploding transformation & (vi-a) OTB & (vi-b) OTB \\ \hline
MSE                     & 712.48              & 361.78                & 493.52                     & 464.07                       & 1785         & 2830         \\
SSIM                    & 0.21                & 0.30                  & 0.23                       & 0.58                         & 0.20         & 0.13         \\ \hline
\end{tabular}
}
\end{table}

Focusing on Figure~\ref{fig:resultsResNet}, the first chart in the first row, (row a), for the CASIA dataset (first two columns) shows that the attacker matching score on the scenario (i) falls below the acceptance threshold (a little above 0.9) from iteration 55 onward even for a high-security threshold (FAR=0.001) and ends nearly at 0.6 at iteration 180. Similarly, for the next four rows on the same column, cancelable biometrics applying Gaussian noise, Laplacian noise, Spread, and Imploding respectively, we can see that despite using these protection methods, the matching score plunges alike almost at iteration 100 below the threshold FAR=0.001. However, this is not the case in the proposed OTB-morph method, (rows f and g), using both LFW and Face mask lite dataset as random biometrics. The output represents that the attacking matching score for the two scenarios of the proposed method plateaued above the threshold FAR=0.001 after iteration 80. While the aforementioned score ends above 0.8 after 180 iterations on scenarios (vi-a and vi-b), it stands at 0.7 at best on scenario (ii) in the end. The proposed method withstands this iterative optimization attack while it offers better performance as well.  If we focus now on the second column, it can be seen that the overlapping area of the impostor and genuine score distributions for the two scenarios of the proposed OTB-morph (scenario vi-a (row d)) is smaller than in the other experimented cases. The same trends are seen for the case of the VGGFace2 (last two columns) although the attacker matching scores slightly go below the threshold for FAR $<$ 0.001 in the case of the proposed method. Considering the first and third columns, the most apparent evolution that can be observed is the falling rate of attacker matching score. While for the first three rows, it decreases drastically to a low Euclidean distance (between 0.6 and 0.7), this pace is far slower for the proposed method, keeping the attacker matching score above 0.8 on both the CASIA and the VGGFace2 datasets. With regard to the score distributions for the VGGFace2 (last column), while the performance drop is not as severe as in the second column, the performance of the proposed method is still better compared to the other scenarios.

Comparing first graph with the third one, in the second row (b), it can be seen that the matching score in VGGFace2 graph falls more than that of CASIA. This finding shows that the CASIA is a bit more robust against the iterative optimization attack than the VGGFace2. The reason behind this difference is that our experimentd ResNet-50 model was pretrained on VGGFace2 datasets. Thus it did better since we picked subjects for morphing from the same dataset. In related works around adversarial samples, this variation is called transferability \cite{ghafourian2023toward}.

For the experiments reported in Figure~\ref{fig:resultsArcFace}, note that the genuine and impostor distributions are more overlapped than before. This happens because our attack framework on Resnet-50 has been built upon TensorFlow in the previous version. Therefore, in order to have the same experimental condition in terms of our threat model, we had to use the TensorFlow implementation of the ArcFace\footnote{https://github.com/peteryuX/arcface-tf2} which is an unofficial version and does not perform as well as Oxford VGGFace implementation used for Resnet-50\footnote{https://github.com/rcmalli/keras-vggface}. Despite the low performance of this ArcFace version, the results follow the same pattern as those of ReNet. The falling rate of the attacker matching score in the first five scenarios for both CASIA and VGGFace2 datasets (first and third columns, rows a, b, c, d, and e) are double that of the proposed method in first 10 iterations. In addition to that, it can be seen in the same graphs that the attacker matching score falls below the threshold FAR=0.1 for other scenarios except for the proposed ones which remained above the EER threshold.

As for the AdaFace, we also had to use an unofficial version\footnote{https://github.com/leondgarse/Keras\_insightface} implemented in TensorFlow but contrary to the ArcFace it performs very well. The corresponding results are depicted in Figure \ref{fig:resultsAdaFace}. From the charts in the first and the third column, it is evident that while the attacker matching score for the proposed scenarios stands above EER, in all other scenarios, it falls either below FAR= 0.001 or oscillates between FAR= 0.01 and FAR= 0.001. Taking into account the performance results of the practiced methods demonstrated in the second and fourth columns in all these figures, the superiority of the proposed method in terms of decreasing the overlapping region of the impostor and genuine score distributions while offering a higher protection rate is noticeable.

Additionally, we reported both Equal Error Rate (EER) and False Rejection Rate (FRR) values, as well as Attack Success Rates (ASR) against the attackers described in Section~\ref{sec3}, for FAR=$\{0.1, 0.01, 0.001\}$ with respect to Resnet-50, ArcFace and AdaFace models in Tables~\ref{tab:results_Resnet}, ~\ref{tab:results_ArcFace}, and~\ref{tab:results_AdaFace} respectively. Starting with Table~\ref{tab:results_Resnet}, we can see that the smallest EER and FRR values are obtained by the proposed method (scenario vi-a and vi-b) whereas the highest value (worst performance) mainly is reported on imploding (scenario v) for both CASIA and VGGFace2. On the other hand, while the Attack Success Rates for spread (scenario iv) at EER and FAR = 0.1 on CASIA are above the values of other scenarios, the first scenario (unprotected biometric system) reported the highest Attack Success Rate in both datasets in overall. Out of the seven scenarios, although Gaussian noise (scenario ii) performed very poor for CASIA at FAR = 0.01 with the corresponding FRR=$67.9\%$, reported FRR results for imploding are worse than all other scenarios in both datasets. Conversely, the proposed method acquired the best performance with FRR= $0.8\%$ at FAR=$0.1$ in both CASIA and VGGFace2.

\begin{table*}[tbh]
\centering
\caption{Comparison of the performance and security of the proposed method using LFW dataset as random biometric (scenario iv-a) and using face mask lite dataset (scenario iv-b) with other scenarios on ResNet-50 model.}
\label{tab:results_Resnet}
\resizebox{\textwidth}{!}{%
\begin{tabular}{|c|cccc|cccc|} 
\hline
\multirow{3}{*}{Scenario} & \multicolumn{4}{c|}{CASIA \cite{22yi2014learning}}                                                                          & \multicolumn{4}{c|}{VGGFace2 \cite{21cao2018vggface2}}                                                                     \\ 
\cline{2-9}
                          & \multicolumn{1}{c|}{\multirow{2}{*}{EER, ASR}} & \multicolumn{3}{c|}{FRR, ASR}                        & \multicolumn{1}{c|}{\multirow{2}{*}{EER, ASR}} & \multicolumn{3}{c|}{FRR, ASR}                      \\ 
\cline{3-5}\cline{7-9}
                          & \multicolumn{1}{c|}{}                          & FAR=0.1         & FAR=0.01         & FAR=0.001       & \multicolumn{1}{c|}{}                          & FAR=0.1        & FAR=0.01       & FAR=0.001        \\ 
\hline
(i)                       & 6.6\%, 88\%                                & 4.6\%, 85.7\%  & 18.9\%, 76.5\%   & 37.8\%, 66\%  & 3.68\%, 90.3\%                                 & 1.8\%, 84.9\%  & 8.7\%, 78.2\%  & 22.4\%, 67.9\%   \\
(ii)                      & 16.6\%, 90.3\%                                & 23.2\%, 86.1\%  & 67.9\%, 12.8\%   & 84\%, 43.9\%  & 8.8\%, 86.3\%                                  & 7.2\%, 85.5\%  & 28.4\%, 71.2\% & 49.4\%, 57.5\%   \\
(iii)                      & 11.29\%, 89\%                                & 12.4\%, 90\%  & 37\%, 75.4\%   & 64\%, 59.7\%  & 5.7\%, 84.5\%                                  & 3.5\%, \_\_  & 17.7\%, 74.7\% & 37.6\%, 63\%   \\
(iv)                      & 14.97\%, 94\%                                & 20\%, 91\%  & 50\%, 76\%   & 74.5\%, 60\% & 8.6\%, 89\%                                  & 7.7\%, \_\_  & 30\%, 76\% & 56.7\%, 63.5\%   \\
(v)                     & 23.6\%, 93.4\%                                & 37.1\%, 86.3\% & 68.2\%, 66.8\% & 86.3\%, 51.4\% & 16.45\%, 88.6\%                                & 22.1\%, 84.2\% & 46\%, 65.7\% & 67.5\%, 50.6\%  \\
(vi-a)                      & 2.69\%, 86\%                                 & 0.8\%, \_\_    & 4.8\%, 71.3\%    & 19.5\%, 28.2\%   & 3.11\%, 86.4\%                                 & 0.8\%, \_\_   & 7.3\%, 73.6\%  & 19.8\%, 41.9\%   \\
(vi-b)                      & 6.29\%, 85\%                                 & 4.3\%, \_\_    & 20.4\%, 57.9\%    & 44.6\%, 23\%   & 6\%, 87.3\%                                 & 4.1\%, \_\_   & 19.3\%, 64.8\%  & 43.8\%, 34.1\%   \\
\hline
\end{tabular}%
}
\end{table*}

In terms of ASR, while the highest percentage on CASIA ($94\%$) belongs to scenario (iv) at EER point, on VGGFace2 it can be seen in scenario (i) at EER with $90.3\%$. Regarding the proposed method, we observed that there is not much difference for ASR at the EER point between all scenarios because the attacker matching score plummets rapidly in first iterations regardless of the protection method. However, the proposed method decreases the falling rate noticeably as the corresponding values for the ASR on scenarios (vi-a) and (vi-b) at the FAR=$0.001$ point are $28.2\%$ and $23\%$ on CASIA and $41.9\%$ and $34.1\%$ on VGGFace2 respectively. The reason we didn't report ASR for FAR=$0.1$ in some scenarios is that the EER is higher than the FRR at FAR=$0.1$.

\begin{table*}[tbh]
\centering
\caption{Comparison of the performance and security of the proposed method using the LFW dataset as random biometric (scenario iv-a) and using the Face Mask Lite dataset (scenario iv-b) with other scenarios on the ArcFace model.}
\label{tab:results_ArcFace}
\resizebox{\textwidth}{!}{%
\begin{tabular}{|c|cccc|cccc|} 
\hline
\multirow{3}{*}{Scenario} & \multicolumn{4}{c|}{CASIA \cite{22yi2014learning}}                                                                          & \multicolumn{4}{c|}{VGGFace2 \cite{21cao2018vggface2}}                                                                     \\ 
\cline{2-9}
                          & \multicolumn{1}{c|}{\multirow{2}{*}{EER, ASR}} & \multicolumn{3}{c|}{FRR, ASR}                        & \multicolumn{1}{c|}{\multirow{2}{*}{EER, ASR}} & \multicolumn{3}{c|}{FRR, ASR}                      \\ 
\cline{3-5}\cline{7-9}
                          & \multicolumn{1}{c|}{}                          & FAR=0.1         & FAR=0.01         & FAR=0.001       & \multicolumn{1}{c|}{}                          & FAR=0.1        & FAR=0.01       & FAR=0.001        \\ 
\hline
(i)                       & 25.79\%, 88.3\%                                & 44.3\%, 70.4\%  & 74.3\%, 24.1\%   & 93\%, 2.5\%  & 18.3\%, 88.2\%                                 & 26.8\%, 78.5\%  & 58.1\%, 39.6\%  & 79.3\%, 13\%   \\
(ii)                      & 32.7\%, 80.1\%                                & 62.8\%, 39.9\%  & 89.5\%, 4.1\%   & 97.6\%, 0.1\%  & 24.7\%, 81.8\%                                  & 44.9\%, 56.6\%  & 77.7\%, 12.6\% & 92.3\%, 1.8\%   \\
(iii)                      & 29.7\%, 82.8\%                                & 54.4\%, 50\%  & 87.8\%, 6.9\%   & 97.2\%, 0.3\% & 22.46\%, 84.3\%                                  & 39\%, 65.7\%  & 74.6\%, 20\% & 89.5\%, 0.4\%   \\
(iv)                      & 30.9\%, 82.4\%                                & 55.3\%, 51.8\%  & 84\%, 12\%   & 94.6\%, 1.7\%  & 23.64\%, 94\%                                  & 41\%, 82\%  & 74\%, 33.4\% & 90.4\%, 10\%   \\
(v)                     & 32.37\%, 87.7\%                                & 56.1\%, 63.4\% & 80.9\%, 16.5\% & 92.2\%, 2.5\% & 24.14\%, 80.2\%                                & 37\%, 62.9\% & 63.8\%, 24.9\% & 79.1\%, 8.6\%  \\
(vi-a)                      & 11.59\%, 19.2\%                                 & 12.3\%, 16.1\%    & 37.5\%, 0.5\%    & 66.1\%, 0.0\%   & 11.3\%, 31.7\%                                 & 12.5\%, 28.4\%   & 37\%, 2.5\%  & 59.5\%, 0.0\%   \\
(vi-b)                      & 14.9\%, 36.4\%                                 & 19.9\%, 25.7\%    & 48.3\%, 1.9\%    & 72.6\%, 0.1\%   & 16.79\%, 48.6\%                                 & 24.5\%, 32.6\%   & 53.4\%, 2.7\%  & 74.4\%, 0.1\%   \\
\hline
\end{tabular}%
}
\end{table*}

Considering Table~\ref{tab:results_ArcFace} we can observe that the results for the performance differ between CASIA and VGGFace2. Although the worst performance is observed in scenario (ii), scenario (v) is almost as inferior as the former. Concerning ASR, the proposed method withstands the iterative optimization attack better while offering higher performance compared to all the other scenarios. In specific, ASR values at the EER point for both scenarios of the proposed method are all below $50\%$: $19.2\%$ and $36.4\%$ on CASIA and $31.7\%$ and $48.6\%$ on VGGFace2. These results are achieved where none of the other scenarios did better than $80\%$ at the same threshold. 

Finally, with respect to the AdaFace, the results convey the same understanding as the previous tables. It can be seen that not only the proposed methods perform best, but also cancelable biometrics generated by this approach are further protective keeping ASR less than $60\%$ at the EER point where it stands above $90\%$ for other scenarios at the corresponding setting. These results show the superiority of OTB-morph compared to related methods both for security protection and recognition performance. 

\begin{table*}[tbh]
\centering
\caption{Comparison of the performance and security of the proposed method using the LFW dataset as random biometric (scenario iv-a) and using the Face Mask Lite dataset (scenario iv-b) with other scenarios on the AdaFace model.}
\label{tab:results_AdaFace}
\resizebox{\textwidth}{!}{%
\begin{tabular}{|c|cccc|cccc|} 
\hline
\multirow{3}{*}{Scenario} & \multicolumn{4}{c|}{CASIA \cite{22yi2014learning}}                                                                          & \multicolumn{4}{c|}{VGGFace2 \cite{21cao2018vggface2}}                                                                     \\ 
\cline{2-9}
                          & \multicolumn{1}{c|}{\multirow{2}{*}{EER, ASR}} & \multicolumn{3}{c|}{FRR, ASR}                        & \multicolumn{1}{c|}{\multirow{2}{*}{EER, ASR}} & \multicolumn{3}{c|}{FRR, ASR}                      \\ 
\cline{3-5}\cline{7-9}
                          & \multicolumn{1}{c|}{}                          & FAR=0.1         & FAR=0.01         & FAR=0.001       & \multicolumn{1}{c|}{}                          & FAR=0.1        & FAR=0.01       & FAR=0.001        \\ 
\hline
(i)                       & 6.65\%, 98.3\%                                & 5.3\%, \_\_  & 12.2\%, 95.6\%   & 22.7\%, 88.6\%       
            & 3.0\%, 98\% & 1.5\%, \_\_  & 4.7\%, 96.6\%  & 8.5\%, 91.5\%   \\

(ii)                      & 14.29\%, 95.6\%                                & 17.1\%, 94.2\%  & 34.6\%, 75.8\%   & 63.28\%, 36\%       
             & 6.07\%, 95.4\% & 4.5\%, \_\_  & 12.9\%, 88.1\% & 29\%, 69.7\%   \\

(iii)                      & 10.04\%, 97\%                                & 10.0\%, 96.9\%  & 24.5\%, 87.2\%   & 46.05\%, 63.2\%        
             & 4.36\%, 96.8\% & 2.7\%, \_\_  & 8.4\%, 92.7\% & 15.8\%, 82.6\%   \\

(iv)                      & 12.95\%, 96.2\%                                & 14.7\%, 95.18\%  & 30.76\%, 82.2\%   & 49.4\%, 55.1\%          
              & 6.24\%, 96\% & 4.8\%, \_\_  & 13.6\%, 88.35\% & 28.6\%, 70.6\%   \\

(v)                     & 21.33\%, 91.6\%                                & 30\%, 83.3\% & 54.8\%, 51.1\% & 71.7\%, 27\%          
              & 13.44\%, 89.82\% & 15.7\%, 87.1\% & 33\%, 65.6\% & 47.1\%, 41.9\%  \\

(vi-a)                      & 1.4\%, 17.3\%                                 & 0.2\%, \_\_    & 1.8\%, 13\%    & 13.3\%, 0.0\%               & 1.09\%, 31.3\% & 0.2\%, \_\_   & 1.1\%, 30\%  & 7.2\%, 3.8\%   \\

(vi-b)                      & 4.1\%, 42\%                                 & 2.1\%, \_\_    & 9.5\%, 16.5\%    & 24.4\%, 1.4\%                & 3.7\%, 59.2\% & 1.9\%, \_\_   & 8.2\%, 36\%  & 19.3\%, 10.35\%   \\
\hline
\end{tabular}%
}
\end{table*}

\subsection{Limitations}
Despite the advantages that the proposed method introduces to biometric template protection methods in terms of higher performance and lower attack success rate against leakage attacks, there are some limitations that need to be taken into account:

\begin{itemize}
 \item The proposed method needs the acquisition of a reference face in each authentication session which is computationally more expensive than those cancelable methods which change the reference once in a while or upon the leakage.
  \item The proposed method might be slower than other cancelable approaches as it imposes morphing in each authentication attempt.
  \item The proposed method needs a random face image to do the morphing in each authentication attempt. This random image can be produced by generative networks. However, our experiments pointed out that the highest protection will be only achieved when the dissimilarity of the new random face is at the highest with respect to comparing to its predecessors. 
\end{itemize}

\section{Conclusions}\label{sec13}

This work has extended our experiments on our introduced cancelable biometrics method which can be categorized as a branch of visual cryptography with the aim of protecting the biometric templates of clients against all kinds of leakage attacks. The original idea is to adopt the concept of the one-time-pad method to biometrics by using random biometrics as auxiliary data in a cancelable biometrics scheme called OTB-morph (One-Time Biometrics via Morphing). To this end, we used morphing as the transformation function to generate an image that embodies two different identities. We then experimented with the proposed idea using a practical implementation for face biometrics. In the reported experiments we used a morphing algorithm based on Dlib and OpenCV for generating the cancelable templates. With respect to previously reported preliminary results \cite{11OTB-Morph}, the present archival paper presents and discusses extended experiments by: increasing the number of iterations for the iterative optimization attacks, using GAN-generated faces as another mean for random biometric generation, and using a pre-trained ArcFace model for additional evaluation. In conclusion, the proposed method improves both the biometric performance and security against the evaluated attacks.

In our future work, our main goal is to investigate how we can maximize the distance between two one-time biometric of the same individual in subsequent sessions to that we can offer the lowest ASR in case of iterative optimization. In specific, we would like to explore how a random biometric with opposite features to the given subject (e.g. ethnicity, skin color, age, sex, and other facial attributes) can help us to meet our goal.

\section*{Acknowledgments}

This work has been supported by projects: PRIMA (H2020-MSCA-ITN-2019-860315), TRESPASS-ETN (H2020-MSCA-ITN-2019-860813), and BBforTAI (PID2021-127641OB-I00 MICINN/FEDER). M.G. is supported by PRIMA and I.S. is supported by an FPI fellowship from Univ. Autonoma de Madrid.

\begin{figure}[h]%
\centering
\includegraphics[width=0.3\textwidth]{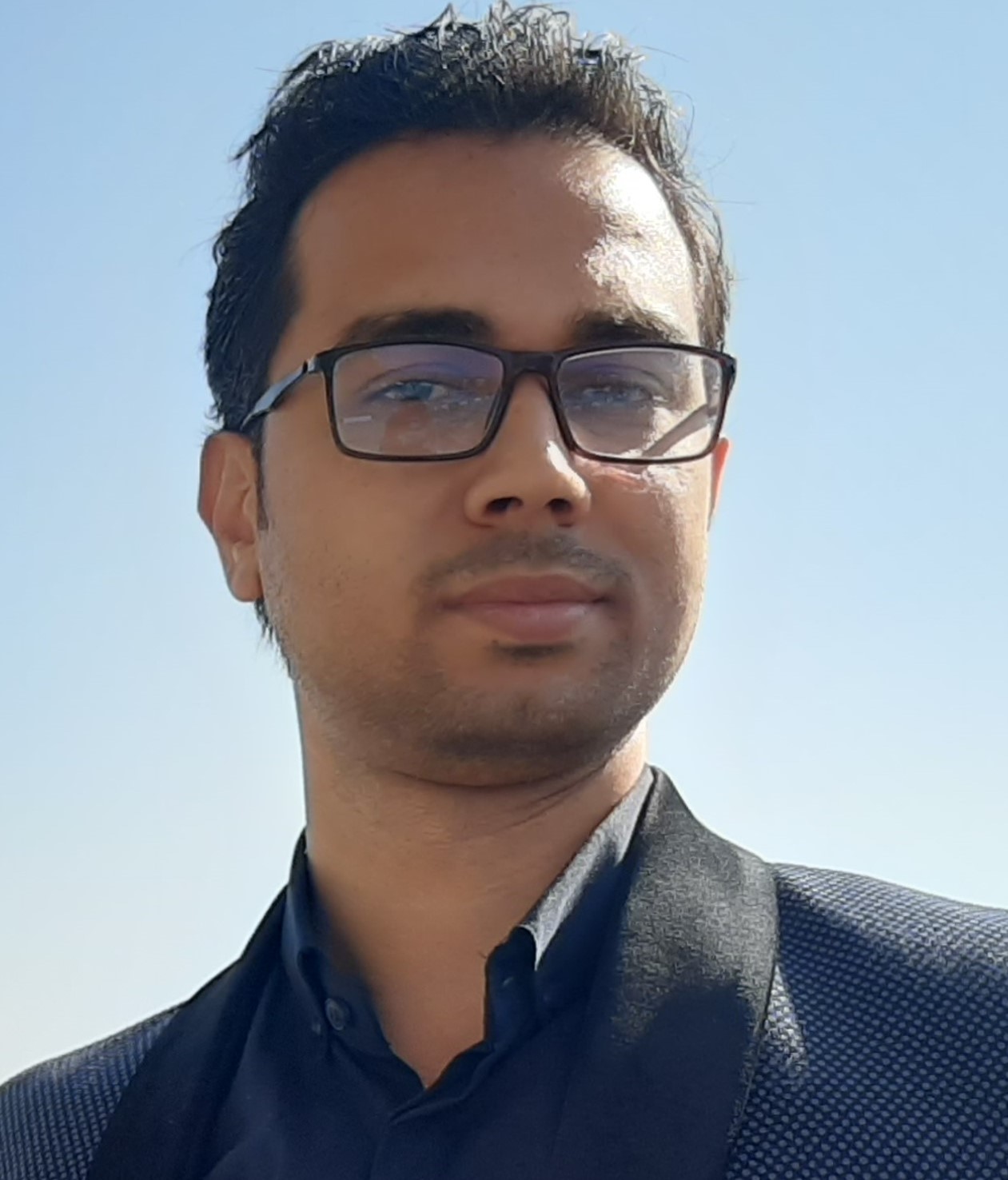}
\end{figure}

\noindent{\bf Mahdi Ghafourian} received his bachelor’s degree in Computer Science from the Islamic Azad University of Mashhad, Mashhad, Iran, in 2011, and his master’s degrees in Information Security and Assurance from the Imam Reza University, Mashhad, Iran, in 2016. He achieved the second position among top Master’s degree graduates. In 2021, he started his Ph.D. with Marie Curie scholarship within the EU ITN project PriMa (Privacy Matters) in the Biometrics and Data Pattern Analytics lab - BiDA-Lab, at the Universidad Autonoma de Madrid, Spain. His research interests include information security, biometrics protection, face recognition, adversarial examples, and federated learning.

E-mail: mahdi.ghafourian@uam.es (Corresponding author)

ORCID iD: 0000-0003-4206-4873


\begin{figure}[h]%
\centering
\includegraphics[width=0.3\textwidth]{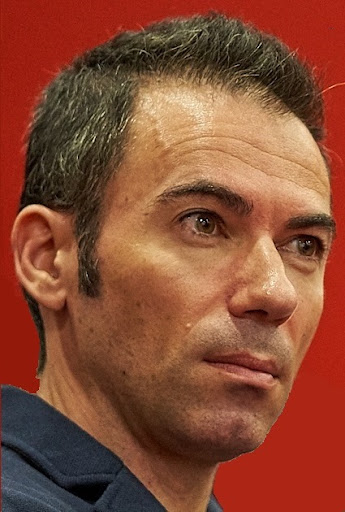}
\end{figure}

\noindent{\bf Julian FIERREZ}\quad received the MSc and the PhD degrees from Universidad Politecnica de Madrid, Spain, in 2001 and 2006, respectively. Since 2004 he is at Universidad Autonoma de Madrid, where he is Associate Professor since 2010. His research is on signal and image processing, AI fundamentals and applications, HCI, forensics, and biometrics for security and human behavior analysis. He is Associate Editor for Information Fusion, IEEE Trans. on Information Forensics and Security, and IEEE Trans. on Image Processing. He has received best papers awards at AVBPA, ICB, IJCB, ICPR, ICPRS, and Pattern Recognition Letters; and several research distinctions, including: EBF European Biometric Industry Award 2006, EURASIP Best PhD Award 2012, Miguel Catalan Award to the Best Researcher under 40 in the Community of Madrid in the general area of Science and Technology, and the IAPR Young Biometrics Investigator Award 2017. Since 2020 he is member of the ELLIS Society.

E-mail: julian.fierrez@uam.es


\begin{figure}[h]%
\centering
\includegraphics[width=0.3\textwidth]{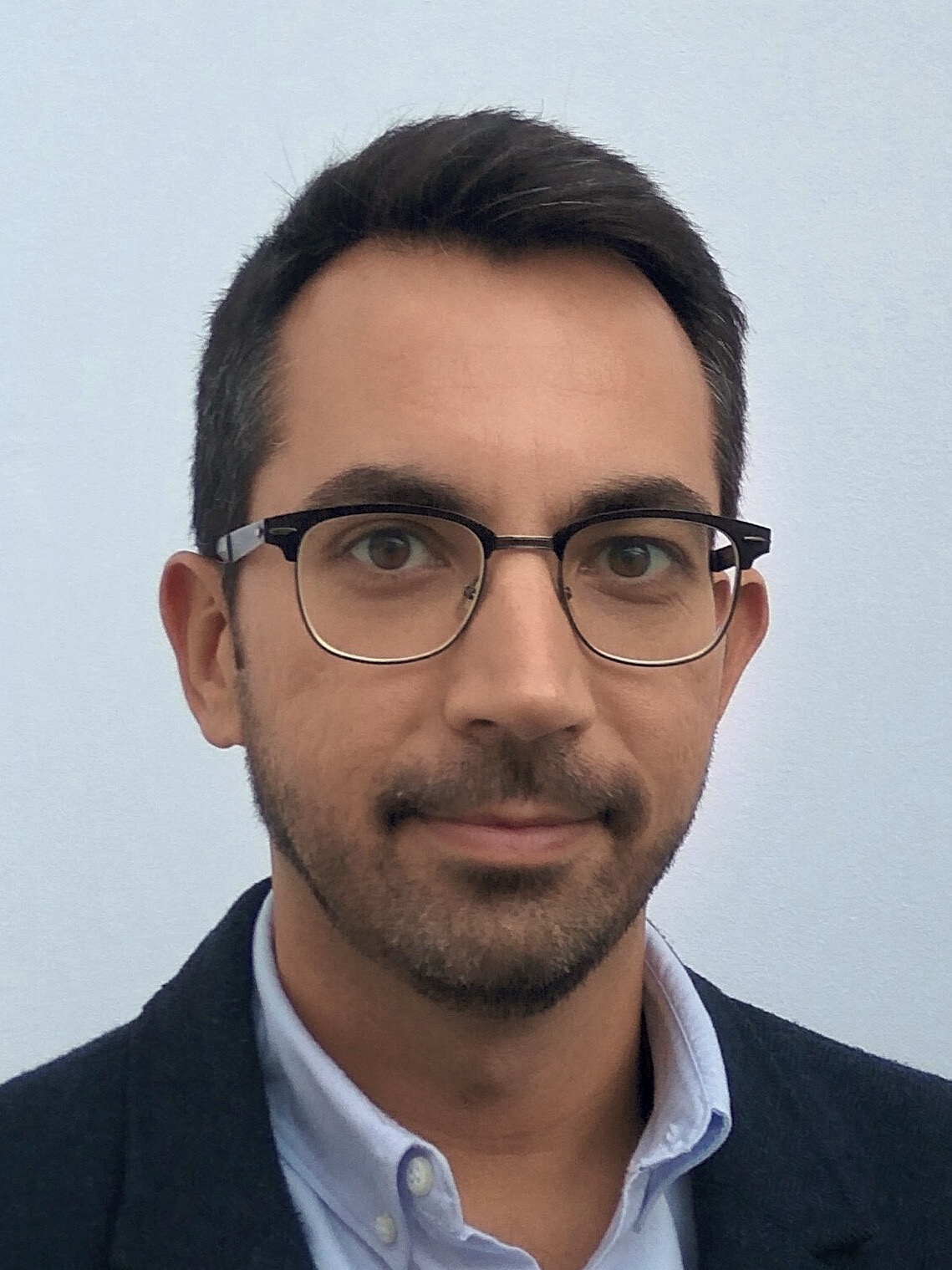}
\end{figure}

\noindent{\bf Ruben Vera-Rodriguez }\quad received the M.Sc. degree in telecommunications engineering from Universidad de Sevilla, Spain, in 2006, and the Ph.D. degree in electrical and electronic engineering from Swansea University, U.K., in 2010. Since 2010, he has been affiliated with the Biometric Recognition Group, Universidad Autonoma de Madrid, Spain, where he is currently an Associate Professor since 2018. His research interests include signal and image processing, pattern recognition, HCI, and biometrics, with emphasis on signature, face, gait verification and forensic applications of biometrics. Ruben has published over 100 scientific articles published in international journals and conferences. He is actively involved in several National and European projects focused on biometrics. Ruben has been Program Chair for the IEEE 51st International Carnahan Conference on Security and Technology (ICCST) in 2017; the 23rd Iberoamerican Congress on Pattern Recognition (CIARP 2018) in 2018; and the International Conference on Biometric Engineering and Applications (ICBEA 2019) in 2019.

E-mail: ruben.vera@uam.es


\begin{figure}[h]%
\centering
\includegraphics[width=0.3\textwidth]{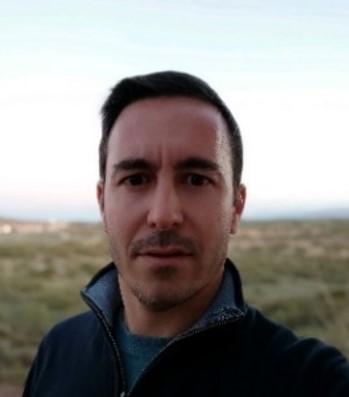}
\end{figure}

\noindent{\bf Aythami Morales Moreno }\quad received M.Sc. in Electrical Engineering, and Ph.D from the Universidad de LPGC in 2006 and 2011. Since 2017, he is Associate Professor with the Universidad Autonoma de Madrid. In his work, he combines his interests in machine learning, biometric processing, security, and privacy.

E-mail: aythami.morales@uam.es


\begin{figure}[h]%
\centering
\includegraphics[width=0.25\textwidth]{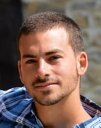}
\end{figure}

\noindent{\bf Ignacio Serna }\quad received B.S. degree in mathematics and B.S. degree in computer science from the Autonomous University of Madrid, Spain, in 2018, and M.S. degree in Artificial Intelligence in 2020. He is currently pursuing a Ph.D. in Computer Science at the BiDA-Lab. His research interests lie in computer vision, pattern recognition and explainable AI, with applications to biometrics.

E-mail: ignacio.serna@uam.es


\end{document}